# The Evolution of Concept-Acquisition based on Developmental Psychology

WEI Hui

Department of Computer Science, Lab of Algorithm for Cognitive Models, Fudan University, Shanghai 200433, P. R. China

**Abstract:** A conceptual system with rich connotation is key to improving the performance of knowledge-based artificial intelligence systems. While a conceptual system, which has abundant concepts and rich semantic relationships, and is developable, evolvable, and adaptable to multi-task environments, its actual construction is not only one of the major challenges of knowledge engineering, but also the fundamental goal of research on knowledge and conceptualization. Finding a new method to represent concepts and construct a conceptual system will therefore greatly improve the performance of many intelligent systems. Fortunately the core of human cognition is a system with relatively complete concepts and a mechanism that ensures the establishment and development of the system. The human conceptual system can not be achieved immediately, but rather must develop gradually. Developmental psychology carefully observes the process of concept acquisition in humans at the behavioral level, and along with cognitive psychology has proposed some rough explanations of those observations. However, due to the lack of research in aspects such as representation, systematic models, algorithm details and realization, many of the results of developmental psychology have not been applied directly to the building of artificial conceptual systems. For example, Karmiloff-Smith's Representation Redescription (RR) supposition reflects a concept-acquisition process that re-describes a lower level representation of a concept to a higher one. This paper is inspired by this developmental psychology viewpoint. We use an object-oriented (OO) approach to re-explain and materialize RR supposition from the formal semantic perspective, because the OO paradigm is a natural way to describe the outside world, and it also has strict grammar regulations. We designed and realized the representations of concept-acquisition and their evolutions from one to another at different levels. Our results solidify the semantics and pragmatics of a concept acquired by RR through a kind of visible, verifiable and replicable carrier—an OO language. Further, in an environment of materializing concepts by objects, an OO language makes the refinement of concepts and the construction of a conceptual system more operable and controllable.

## 1. Introduction

The construction of an expert system entails two main problems: a bottleneck of knowledge acquisition and the domain-restricted application of a knowledge system[1]. The former involves how to translate knowledge of the real world into knowledge that the expert system can use, while the latter involves how only an expert system can adapt to domain-restricted problem solving, because if the issues related to a system are domain-opened, then the system's capacity for problem solving will be very vulnerable. The main reason for these problems is that expert system knowledge is largely mechanical and lacks of the support of underlying semantics. Conceptual systems are not constructed from the perspective of concept development. Designers of intelligent machines believe that conceptualization, conceptual systems, and the formalization of conceptual systems are different. Further, this difference is based on a profound philosophical question: What is the "grasping" of a concept? The researchers do not believe, for example, that a machine is wise enough to grasp a concept when the conceptual structure made by its designers is coded into its memory components [2].

Knowledge acquisition and knowledge representation are the foundation of knowledge-based systems. How to acquire knowledge from the outside world and how to utilize the acquired knowledge efficiently has great significance in the constructing of a knowledge-based system. Aim at this situation, many researchers want to improve the performance of intelligent systems from the respects of knowledge acquisition and knowledge representations [3]. Other researchers aim to improve the performance of expert systems by improving the efficiency of the knowledge acquisition algorithm in the process of inductive learning [4]. However, these methods do not consider the development of human cognition, and as such can not be used to construct a "human-level" knowledge-based system. Because knowledge is an important foundation of human cognitive skills, many artificial intelligence researchers have carried out analyses of the structure of knowledge. For example, one study explained the structure of human knowledge from different angles by classifying and summarizing various types of human knowledge that employs different strategies of knowledge acquisition and knowledge representation for different types of knowledge[5].

Other research has analyzed the special structure of expert knowledge by extracting and representing expert tacit knowledge [6]. Again, however, such studies did not take into account the psychological model of humans, and thus their results can not fundamentally improve the sharing and flexibility of knowledge in an intelligent system Although the problem solving ability of human beings is related to their knowledge structure, very little research has attempted to explain this relationship clearly [7]. Further, research on the representation and development of conceptual structures has been receiving more attention recently. The goal of such research is to construct knowledge structures and broaden the base of problem solving. The representational methods used in this area of research include not only classical knowledge representation, such as semantic networks and frameworks, but also involve newer methods such as Conceptual Graphs[8,9,10], Ontology[11,12,13], and Concept Lattices[14,15,16,17]. However, these methods are only suitable for knowledge that has a good structure, and they do not consider the process of conceptual formation. That is, the problem solving served by these methods centers around true/false and here/what/who/which questions. Therefore, the distance between the semantic form of signs and conceptual connotation is great. This paper is based on the research results of cognitive and developmental psychology. We aim to construct a conceptual system that can enrich itself continually, using the knowledge it has acquired to carry out various types of problem solving wisely.

To solve various problems that exist in traditional artificial intelligence, the construction of AGI (artificial general intelligence) systems have become a trend in artificial intelligence research, such as the AGIRI research team led by Ben Goertzel [18]. In addition, a conference related AGI will be held at the University of Memphis in the United States in 2008[19]. The goal of AGI is the development of a software system that has extensive and diverse intellectual function, or "human level" intelligence, including the ability to understand various things, the ability to solve new problems and the ability to use rich language for communication [20]. An AGI system with large amounts of knowledge and skills is insufficient; it should have the ability to improve its cognitive skills through learning and apply its cognitive skills flexibly to problem solving. To achieve AGI, researchers have put forward many solutions, such as John R Anderson's ACT-R cognitive structure [21] and the Cassimatics NL's Polyscheme cognitive structure[22]. ACT-R researchers aim to find out how humans organize knowledge and generate intelligent behavior. They believe that ACT-R can eventually perform most, or even all, human cognitive skills with sufficient research progress. Polyscheme solutions, on the other hand, imitate human intelligence by various methods, such as knowledge representation, problem solving, and reasoning. Polyscheme researchers believe that as long as the system can solve the same problems and perform the same reasoning as humans, then the system can achieve "human level" intelligence, regardless of whether the mechanism is similar to the human brain. Finally, in the NARS system, proposed by Pei Wang[23], the main difference between it and traditional reasoning systems is that the NARS system has the ability to learn from past experience. It attempts to provide a unified theoretical model for AI, with the ultimate goal of this research being to create a "thinking machine". This paper, based on the current situation of the construction and application of expert systems in the field of artificial intelligence being domain-restricted, assumes that due to a lack of rich semantic support, current expert systems have problems in the process of acquiring knowledge and solving problems. These problems are also the fundamental reason that other knowledge-based systems have low performance. Thus, the construction of a complete conceptual system has great significance with respect to the solving of unrestrictive problems. It can be seen from above, the current research and AGI have many common features: each has the aim of creating intelligent systems that can carry out domain-unrestrictive problem solving, and each believes that intelligent systems have the ability to learn knowledge automatically and use the knowledge flexibly.

Artificial intelligence methods usually are not concerned with how intelligent behavior was achieved by biological prototypes (e.g., humans) in the field of computer science. That is to say, the physiological and psychological mechanisms of biological prototypes are not used as a reference as long as intelligent behavior can be achieved. However, psychologists tend to use well-designed experiments to outline theoretical models of cognitive process when they study the cognitive skills of humans. These theoretical models often include high level processing, and try to explain all phenomena observed from subjects. Because such models are very complex, it is difficult to express all the details of their realization. This results in artificial intelligence methods and psychological models converging in a way that prevents the former from consulting the latter. For example, on the one hand, there are many concept representation methods in the field of artificial intelligence, but these methods also have obvious limitations; on the other hand, psychologists have proposed macroscopic ideas for concept acquisition, but they have ignored the details of realization. In this paper, we attempt to find an abstract level of concept representation at which we can integrate the ideas of the two fields together. In the fields of cognitive psychology and developmental psychology, the representation and development of conceptual systems have long been studied. Furthermore, psychology is more forward-looking than computer science with respect to the research of concepts.

The Representational Redescription (RR) model proposed by Karmiloff-Smith is a very promising hypothesis[24] that depicts a complete outline of cognitive development. The model has four representational levels and claims that people acquire knowledge through a proper Representational Redescription progress. However, because of the difference on research purpose and method between psychology and AI, RR model still have many gaps and faults. Many details and mechanisms of realization have yet not to be considered, so the RR theory still incomplete. Because the theory is still very vague, this paper proposes a multi-level, object-oriented representation approach for representing and constructing conceptual systems. It uses object-oriented language syntax specifications as a means of semantic formalization, and depicts the process of concept representation and development on the levels of Implicit (I), Explicit 1 (E1), Explicit 2 (E2) and Explicit 3 (E3). It provides a detailed scheme for realizing the RR hypothesis, including the specific form of the four representation levels and three change phases. Furthermore, it depicts the process of concept acquisition and cognitive development through changes in representation form. To this end, it helps to know more about representational forms and the principles related to using conceptual systems.

This research combines computer science with psychology. First, it adopts a new perspective of concept development in the field of psychology. It explores some vague assumptions of psychology from the computational view in the process of realizing representations. Therefore, to a certain extent it complements psychological research in realizing mechanisms. Second, this study follows the main principle of constructing conceptual systems, i.e., to broaden basic knowledge, which is very different from traditional expert systems. It also provides the possibility of solving unrestrictive problems for conceptual systems, and it also helps to improve the reasoning ability and problem solving ability of knowledge-based systems. The objective of the research is to promote the construction of knowledge systems and unrestrictive problem solving in the field of artificial intelligence, while also complementing psychological research that currently lacks necessary details related to system realization and concrete mechanisms.

The organization of this paper is as follows. Section 2 introduces the significance of a conceptual system that has rich semantic elements. Section 3 introduces the RR hypothesis and its enlightenment to concept acquisition and concept representation. Section 4 outlines the process of object-oriented Representational Redescription. Section 5 describes the evolution of representational forms and the promotion of problem solving. Section 6 is the conclusion. Because of limited space, this paper only discusses the forms of representation and is not concerned with generating algorithms of these representational forms. Such algorithms require the support of conceptual hierarchical structures and inductive learning.

## 2. A conceptual system with rich semantic elements is the key to intelligence

### 2.1 Practical problem solving requires abundant knowledge

Figure 1 (a), (b) and (c) show examples of problem solving carried out by kindergarten children. The children were asked, according to the content of the given picture, to either match a T-shirt with trousers, locate a ball the same size as another, or figure out which was the longest pencil in a group. To accomplish these tasks, the children needed to have learned some basic concepts or knowledge about clothes, toys, and stationery. If a problem is presented to an intelligent system in the form of a picture, to obtain an answer will require image understanding and knowledge-based reasoning. Realizing this in an artificial intelligence system, however, is quite difficult. Even if an intelligent system can solve one such problem, when we ask it to adapt to all three problem types, will it be able to produce an answer? The three examples above are just a minimal sampling of the intelligent tasks that children can accomplish easily. The multi-tasking of problem solving in children is a great challenge to intelligent machines. Take Figure 1 (b) and (c) as examples, they all involve measuring length, so how does the AI system know the distance between which the two pixels should be measured when calculating the diameter shown in Figure.1 (b)? And how does it know the distance between which two pixels should be measured when calculating the length of the longest pencil in Figure 1 (c)? When these rules are well defined, for example, if when facing a problem that estimates whether a giraffe's neck is long enough to be able to return a bird back to its nest in a tree, how is this rule of measuring length represented?

These examples of problem solving may seem very simple, and indeed they are all solvable by preschoolers. On the surface, such a problem is less difficult than solving an indefinite integral, but it actually reflects one of the big differences between special systems and general systems in terms of quantity and flexibility of knowledge. Designing programs for solving indefinite integrals is currently only an engineering problem, but developing an intelligent system that can solve all the problems illustrated in Figure 1 is still a strongly theoretical issue that has not yet to be solved. In short, these few "simple" examples have offer profound insights into studying the theory of knowledge-based systems. Although these problems that five-year-old children can answer quickly are very simple, and do not belong in the scope of application that artificial intelligence should consider, they do reflect the character of the human cognitive system.

This raises several questions. First, while recognizing the premise that knowledge is important for knowledge-based systems, why is it necessary to also emphasize the dependence of concepts and their developmental processes? Can we use the classical " hypothesis of low-level structure discontiguous" to avoid this complicated problem? Second, if it can not be avoided, then how do we construct the conceptual system and adapt it to multi-tasking? Third, can the construction and development of conceptual systems be formalized? We attempt to answer these questions below.

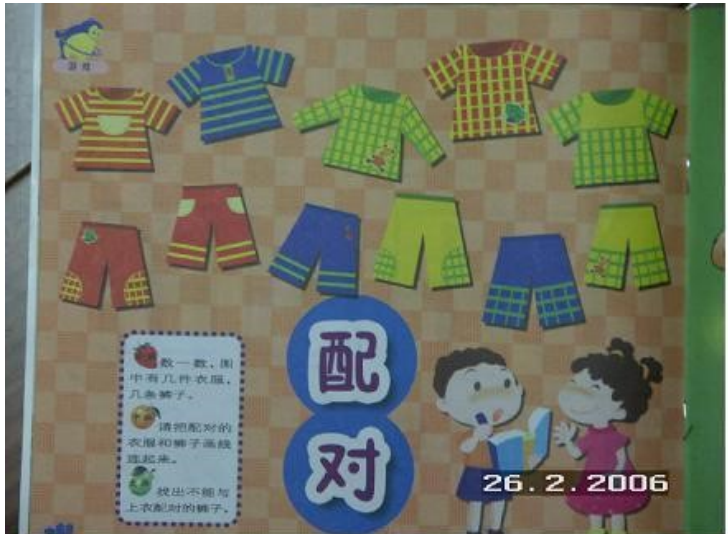

(a) The problem solving task here is to match each jacket with its appropriate trousers.

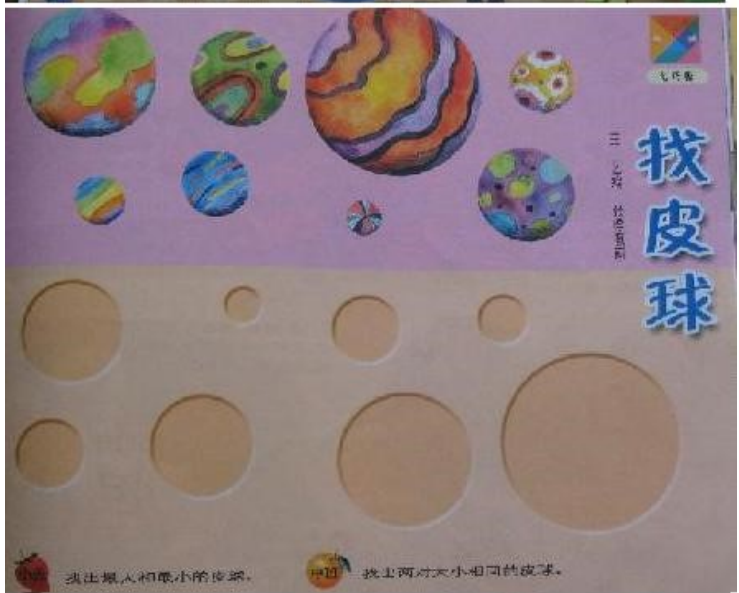

(b) The problem solving task here is to locate the biggest and smallest balls, as well as those with the same size, and match each ball with the right hole.

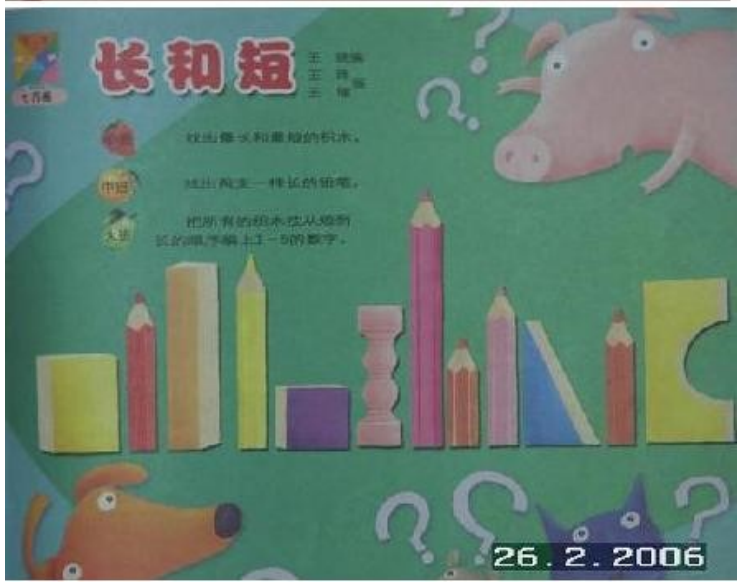

(c) The problem solving task here is to locate the longest pencil and toy brick.

Figure.1　Examples of children's problem solving tasks in different fields

## 2.2 A conceptual system with rich semantic elements helps promote the performance of intelligent systems

Humans exhibit skillful cognitive behavior, such as problem solving, reasoning, decision-making, programming, natural language understanding and scene understanding; all of these abilities depend on a comparatively complete conceptual system[25]. Such behavior is different from that of artificial intelligence systems designed for special applications. Such systems usually have strict restrictions in problem content, representation format, application background and predetermined conditions, while the conceptual system of humans can cope with domain-open problems. That is to say, for coping with such a broad number of uncertain requirements, the knowledge system of humans must be general rather than special. Humans use a single system to solve all kinds of intelligent tasks, not just to perform a single certain task. The use of unrestrictive knowledge requires a complete conceptual system[26], for which there are high demands in the required amounts of knowledge and the representation, storing and processing of knowledge. The human conceptual system is very complete, and can meet the needs of solving different types of problems. This is precisely what current AI systems lack, and as such a human-like conceptual system is one of the most important research directions of the field of AI.

Most knowledge representation research incorporates domain-restricted knowledge-based systems that are strictly limited by the tasks, reasoning abilities, and architecture available from current computers. As such, most current AI systems[27,28] belong to a certain specific area, and thus these systems are only able to adapt to simple, well-structured knowledge or rules. In the face of other areas that require a lot of background knowledge and common sense, however, the results produced by such systems is insufficient. In the real world, problem solving is carried out by people usually across several areas; for example, children learn the concept of support when they play with toy bricks, and it is at this time that the concept of support is linked with the toy bricks. Then, the children may encounter a new problem, such as how to help ants cross a river, in which case they might decide to use some tree branches as a bridge. This solution shows the application of the concept of support to the river bank and branches. Such application shows that children have really grasped the concept of support, embodying real generalization when solving problems.

Viewed from the knowledge engineering point of view, the above several examples illustrate that common problem solving often requires knowledge of other areas or common sense. However, such knowledge still poses great difficulties in terms of both acquisition and representation, which have led to the current problem of AI systems not being able to carry out problem solving that crosses several areas. Take the concept of "apple", for example; expert systems in different areas, such as plant taxonomy, nutrition, horticulture and genetics, focus on different attributes. The concept of "weather" is another example, with different time limits yielding totally different types of answers, such as today's weather, the weather one month ago, the weather 300 years ago and the weather 100,000 years ago. These two examples demonstrate the conceptual operation for the purpose of specific application is likely to cause separation within the definitions of knowledge. If the conceptual structure system is poor, then the application of knowledge will be limited.

Therefore, the issue of how to break through such limitations is very important and needs to be resolved urgently[29,30].

**2.3 Concept acquisition is a developmental process**

The construction of a conceptual system can not be accomplished in a single action. It must go through a developmental process, forming gradually by uninterrupted learning. The conceptual system of humans evolves from infancy, and the character of this process is progressive, hierarchical, continuous and dynamic. It is an uninterrupted process similar to cultivating crystals in a solution. Many previous studies on the processes of human cognitive development have been done in the field of psychology[24,31,32]. Because of the crucial role of a conceptual system in cognitive processes, its structure, representation and construction become key to understanding the knowledge system of humans. However, the "hypothesis of low-level structure discontiguous" of knowledge and conceptualization of classical AI is different from the conceptual system of human beings. The former tends to only consider special knowledge, and its main principle is "insufficient to supply again". Modern new developments in natural language understanding, scene monitoring, real vision, web information retrieval and video retrieval, however, have prompted us to re-examine carefully why humans can become so versatile. What elements are behind their multi-tasking abilities? The answer to this is, simply, their conceptual system. A lot of research on conceptual systems has applied knowledge of psychology to the field of AI, and achieved some promising results. Some studies, for example, have combined psychology with the field of neural networks to construct new neural networks that can acquire more complex concepts[33,34].

## 3. RR hypothesis and its enlightenment to concept acquisition and representation

**3.1 Outline of RR**

A number of investigations related to the RR hypothesis have been reported in the field of developmental psychology. Two representative views are Piaget's cognitive constructivism and Fodor's nativism. The former believe that human cognition is acquired, but the latter believe that innate cognitive modules play a decisive role in the process of cognition. Integrating some aspects of the two theories, Karmiloff-Smith has proposed the Representational Redescription (RR) model[24]. She believes that humans acquire knowledge through a proper Representational Redescription progress, and that the same knowledge can be stored in various forms and levels. The whole process consists of three phases and four different levels of representation.

RR includes the following cycle. First, information emerges in applications that have special purposes and independent functions. Through the RR process, the information can then be gradually used by other parts of the cognitive system. In other words, RR is a process in which implicit information gradually becomes explicit information. RR is the same in each field, occurring circularly throughout the entire process of cognitive development.

RR supposes that the representation and re-representation of knowledge at least includes four levels, and that different levels have different representational forms. The four levels are: Implicit (I), Explicit 1 (E1),

Explicit 2 (E2) and Explicit 3 (E3). These different forms of representation are not related to age. They occur repeatedly in different micro-fields throughout the entire process of cognitive development. The following are the characteristics of each level, described in psychological language.

**Implicit I:** This form of representation is the module that reflects and analyses the outside stimulus. The stimulus is stored as an independent instance, and behavior is holistic. Representational relations between different fields can not be formed at this level, and thus the presentation of this level is not flexible. In this representational level, information is coded in the form of a program that includes a sequence of atomic operations.

**Explicit E1:** Several representational instances of Implicit I are induced in this level. A flexible cognitive system starts to form gradually, and this establishes a child's immature theory. These E1 representations can be operated on and put into contact with other representations that have been redescribed. At this level, the representational form is acquired by re-coding and compressing the Implicit I program. It is worth noting that the former representations of Implicit I are still stored intact in the minds of the children, and they can still be adapted to certain specific cognitive goals. Although the representation of E1 can be used as systemic material, it can not be accessed through consciousness and be reported in the form of language. RR assumes that only by surpassing the E1 level can a representation be accessed through consciousness and be reported in the form of language. Access through consciousness requires the redescription of the E1 presentations so that **Explicit E2** can appear. Explicit E2 has similar coding to E1.

**Explicit E3:** Representation at this level can not only be accessed through consciousness, but can also be reported in the form of language; knowledge is re-coded into a new form that can cross different systems. This general form is very close to natural language, and it is an easily translatable stable form that can be used for communicating, which is the highest level of RR.

### 3.2 Investigating the RR hypothesis from the perspective of AI

From the perspective of AI, the connotation of RR in knowledge-based systems and machine learning, areas that are not the concerns of psychologists, can be discovered. Viewed from its own objective, the RR hypothesis tries to illustrate that children represent concepts by phases in the process of acquiring concepts, and that the representational forms of those concepts will gradually change and become flexible and reusable. From the point of view of intelligent behavior, RR investigates how the human ability for acquiring and using knowledge evolves, which is consistent with the long-term objective of AI, namely, that AI systems can first acquire some knowledge, but then surpass the level of simply repeating the acquired knowledge, while also having a certain degree of extensible and adaptive capacity in behavior. In short, the RR hypothesis is consistent with the viewpoints of information processing. It is first concerned with the original forms of conceptual incidents. Then, it changes the information into abstract knowledge through learning and inner processing. Finally, the abstract knowledge is memorized to prepare for solving problem flexibly in the future. However, we also find that the RR hypothesis is still far from meeting the requirements of AI. The RR hypothesis only illustrates the standard of estimation and the objective that should be achieved; it does not involve any methods of realization, which are indispensable in AI. These limitations will be discussed in detail in the next section.

The RR hypothesis provides a new idea for constructing a knowledge-based system. It illustrates the development of acquiring and using concepts from the viewpoint of representation evolution. The RR hypothesis describes the process of conceptual development reasonably, that is, it use changes in representation to indicate the improvement of intelligent behavior. The RR hypothesis plays an important role in simulating the cognitive structure of humans from the viewpoint of algorithms.

### 3.3 Shortcomings of RR in constructing a knowledge-based system

There are some shortcomings in the RR hypothesis. It lacks some basic explanations, such as "What is representation? "How is it achieved?" and "How do representations evolve?" Even the difference between "implicit" and "explicit" is still very vague.

The core of RR is the distinction between implicit knowledge and explicit knowledge[35]. However, the RR hypothesis does not elaborate on the specific form of the knowledge in each level. The other problem worthy of note here is that the standard used by RR is very vague. For example, the RR hypothesis rarely describes the representation of E3, and does not clearly illustrate the distinction between explicit E2 and explicit E3; indeed, the single phrase "very close to natural language, it is easy to communicate" is unconvincing. What is the representation of each explicit level? What are the distinctions between them? The RR hypothesis does not give a detailed description. Another similar example is the phrase "accessible through consciousness". Moreover, the RR model includes four representational levels and three change phases, and so there must be a complex mechanism to control the process of RR. But none of these details are included in the RR hypothesis.

To sum up, if you want to apply the RR hypothesis to constructing a knowledge-based system, the questions of how to represent the knowledge of different levels, and which features should be chosen and in what form should they be redescribed still need to be solved.

## 4. RR process based on Object-Oriented standards

### 4.1 Why we chose object-oriented technology to formalize representation

This section describes how to redefine the RR process from the view of AI while keeping it consistent with the construction of a knowledge-based system. Because the RR hypothesis does not include the details of psychological realization, we use the object-oriented method to realize the representation of each level.

The RR model describes changes in terms the previously described four representation levels. It is concerned more about changes in content rather than changes in form. Structure determines function, and the improvement of concept application indicates that the representation has changed. That is to say, both the structure and form change in the model. These changes can be simulated by algorithms. In this paper, we use the object-oriented approach for the construction and development of a knowledge-based system. Because objects are used to describe the attributes and behaviors of concepts. Object-oriented technology has many good features, such as modularity, maintainability, extensibility, and inheritability. This is why we combine object-oriented technology with knowledge representation. Moreover, object-oriented technology is a good tool for describing the physical world. It has a reliable guarantee in both theory and methodology. Taking the learning of the concept of "counting", a very important cognitive skill, as an example, this paper illustrates the RR process of acquiring the concept in the following sections. The process uses changes in the attributes and behaviors of an object to realize the evolution of mastering the concept, and to explain the development of cognition.

**4.2　Acquisition process of the concept of "counting"**

"Counting" as an important and basic cognitive skill in the process of the growing up of children that has been studied by many psychologists. Gelman and Gallistel proved that the early "counting" of children is not only a mechanical learning process[36]. Although children may make mistakes when they learn how to count, their behaviors are constrained by the principles of counting. The first principle is one-to-one matching. It grasped this fact: people must match each object in one set to only one object in another set, and then to determine whether the sum of the two sets are equal. Children may make mistakes when they try to count, but they rarely violate the principle of one-to-one matching. They use a unique figure to mark each object of the set to be counted. The second principle is the steady ordering of natural numbers. For example, when a child wants to count four objects, he may count "1, 3, 7, 10" as long as these figures are different and their order is the same, although the "counting" is very peculiar. Gelman believes that children's counting accords with the relevant constraints of natural numbers. In addition, there are three other principles that restrict children's counting: the object-unrelated principle, the sequence-unrelated principle, and the cardinal sum principle. The object-unrelated principle and sequence-unrelated principle stipulate that any type of object can be counted, while the cardinal sum principle is unrelated to the order of counting; that is to say, we can count a row of objects from its beginning or its middle, as long as each object is counted only once. The cardinal sum principle stipulates that only the last counting number represents the cardinal sum. In the following realization of representation, we take "counting" as an important example for describing the representational form of each level.

**4.3 Using the RR hypothesis to re-explain the "counting" process and using object-oriented standards to represent it**

In order to describe the development of the counting process more clearly, here we offer some additional explanations before realizing the formalization of each representational level. Using the RR hypothesis to re-explain how children acquire the concept of "counting", we include some observable and measurable representative behaviors. These behaviors require realization through formalizing presentation.

Table 1 describes the typical behaviors involved in mastering the concept of "counting" in the Implicit I

level, the underlying characteristics of representation, and the usable formalized strategy.

| | Typical behaviors of Implicit I | Explanation of behaviors from inner representation | Strategy of formalization |
|---|---|---|---|
| 1 | At this level, children only learn some typical instances of counting, such as counting apples, toy bricks and candy. | Many event instances are saved independently; even though they are very similar, they do not share representational components, and so they are not as abstract as concepts. | Objects belonging to the same class are formalized. Each object is coded independently. At this time the coding records the events as a whole, and keeps the detailed parameters related to when the events happened. |
| 2 | The children's counting behavior is not flexible. They can only repeat the actions that they have learnt. When receiving the instruction to count the apples, they mechanically point to each object one by one, and say a number to match each. They repeat the last number as the answer to reply the questioner, but in fact they do not understand that the last counting number indicates the cardinal sum of the set to be counted. | Representation is composed by a series of basic atomic operations. It is a true record of the counting process. It is only a reaction response to the specific situation or problem. | We use atomic operations to indicate fundamental behavior. The whole counting process is composed of a series of atomic operations, and the order of these atomic operations is fixed. These atomic operations constitute a continuous unity, but they have no modularity. |
| 3 | At this level, behavior forms a unity that can not be divided. Each counting incidence of the children involves carrying out the whole process of counting. | The representation only has one interface to the outside. It has not been divided into independent functional submodules, according to the principles of function reapplication and function separation. It can not be shared, and so it can not be connected to any outside events or concepts. | The constraints of all attributes and operations in an object are "private"; namely, these attributes and operations have no external visibility, so they can not be accessed or used by other parts of the conceptual system. The programs have only one entrance and one exit |
| 4 | The children can only repeat the same counting behavior under the same conditions; they can not apply the counting behavior to a new environment. Because of this lack of flexibility in counting, they can count neither rearranged apples nor newly appeared beans. | The representation, which still does not realize the essential characteristics of counting behavior, has many attributes that have not been divided. For example, the variation of the counted objects is still unknown; the counted objects can not be extended to other fields. Redundant attributes probably still exist. | All attributes of instance are constant, and the parameters of the atomic operations are also constant. Thus, the whole instance can be considered as a set of constants. |
| 5 | At this level, several mistakes might occur during the counting process, such as the counting number said in response not matching the object the children point to. | Some components of RR are not fully matured, such as the ordinal concept not being completely formed. Thus, their reliability can not be guaranteed by drawing support from the other mature components. | The reference to mature objects are never or rarely included in the representation. |

Table. 1 Behavioral characteristics of implicit I and the strategy for realizing representation

```
Instance    Counting Apples        /* master the behavior by counting apples or observing others count apples*/
{
    private :      /* "private" indicates that the following attributes can not be seen from the outside, because it is implicit*/
```

```
    const Sound    ONE, TWO, THREE;    /*constant of voice*/
    const Person   ME;                  /*constant of role*/
    const Room     ROOM1;               /*constant of site*/
    const Table    TABLE1;              /*constant of properties*/
    const Apple    APPLE1, APPLE2, APPLE3;   /*constant of properties*/
    /*extracting the common components, some are essential attributes, while some are occasional components*/
  private:    /*implicit representation can not be seen*/
    In(ME, ROOM1);
    On(APPLE1, TABLE1);
    On(APPLE2, TABLE1);
    On(APPLE3, TABLE1);
    InLine(APPLE1, APPLE2, APPLE3); /*arrange in line */
  /*the following is the real memory of some actions; here, there are some atomic operations*/
    ME.Move(HAND);
    ME.PointTo(APPLE1);
    ME.Say(ONE);

    ME.Move(HAND);
    ME.PointTo(APPLE2);
    ME.Say(TWO);

    ME.Move(HAND);
    ME.PointTo(APPLE3);
    ME.Say(THREE);

    ME.Say(THREE);
}
```

Table 2 describes the typical behaviors involved in mastering the concept of "counting" in the Explicit E1 level, the inner characteristics of representation, and the usable strategy of formalization.

| | Typical behaviors of Explicit E1 | Explanation of behaviors from inner representation | Strategy of formalization |
|---|---|---|---|
| 1 | At the E1 level, children have already gained much experience in counting apples, so they can not only repeat the counting instances they have learnt, but also implement new counting behavior accurately. Moreover, similar parts can be observed from the behavior. | By summarizing counting examples, children discover the common components of counting behaviors and form common representations, which still require further evolution. | We use class to indicate the counting concept we have gained. The codes inside this class can be shared by examples of all other classes. The class is the abstract representation of the counting behavior, and each counting behavior is an instance of the class. |
| 2 | Children have the ability to count using objects. When asked questions such as "How many apples are there?", they can point to each apple one by one and respond with the number '1,2,3…" to count the apples. Finally, they repeat the last counting number as the answer to the question. | The representation, as an independent module with a relatively single function, is considered the core of counting behavior. The module contains characteristics of counting behavior, such as circulation, and the ability to return the counting result. The representation can be applied to some other, similar types of problem solving. | We use a member function, considered to be the typical behavior of class to indicate the entirety of counting behavior. Inside of the function, there are circulation structures and other, smaller subfunctional parts that return the counting result. The function also contains some variables, and so its adaptability can be expanded to a broader scope. Being the external interface of class, this member function can be called in other situations that require counting. |
| 3 | At this level, some counting principles become explicit, such as the principle of one-by-one matching. Children will match each apple with a number in an increasing order, and neither miss any apple nor count one apple twice. They can count the apples accurately in different arrangements and quantity. When they point to the objects, their fingers move accordingly in linear order so as to avoid missing one. | The principle of one-by-one matching can be found in representation: matching the apples with natural numbers in an incremental order. Therefore, two closely connected function modules will appear in representation. Namely, the linearly scanned modules, which neglect the current arrangement of apples, and the modules that can match the number with each apple. | We use "app_set" to represent the set of apples so as to ignore their arrangement and order. The function "CountingApples()" contains two small modules: the "Index (app.set)" module and the "OneToOneMap" module. |
| 4 | Children begin to understand the nature of counting, knowing that the sum of apples is irrelevant to the person who is counting the apples and the environment where the counting behavior takes place. Therefore, their counting behavior has a certain flexibility, such as "counting the apples on a small table", meaning that they can carry out counting under different environments. | By removing the environment attributes that are irrelevant to counting behavior, the generalization of the representation is enhanced. In addition, the representation becomes more abstract, and it can be adapted to more situations. It can breakthrough the restriction of teaching models, and more easily connect with objects from new fields. | The attributes of counting class become variable. The external visibility of member functions, which is used to record typical behavior, changes from "private" to "protected". Thus, these functions can be shared with the outside. |
| 5 | The principle of cardinal sum becomes explicit. When asked to tell the sum of the objects in the same group that has been counted before, children report the last counting number as the cardinal sum instead of counting them again. | Through inducing many instances of counting behavior, some new attributes appear in the representation such as the final outcomes of counting. | We define a new attribute "result" in order to preserve the counting result when the counting process ends, and prepare for being used in the future. |

Table 2 Behavioral characteristics of explicit E1 and the strategy for realizing representation

```
Class    CountingApples    /*Instances were changed into a general class through inducing, but it is still limited to
                            the scope of "counting apples"*/
{
```

```
private :    /*some attributes were parameterized, so it has a certain flexibility*/

   const      numlist;      /*ordinals are determined, and the order of numbers becomes important*/

   Person     p;            /*who is counting becomes unimportant, so it is represented by a variable*/

   APP_Set    app_set;      /*amount and the laid shape of apples becomes unimportant, so it is represented by a set,
                              and is also a variable*/

   int        result;       /*a new attribute, it is used for saving the counting result; the sum of apples can be
                              accessed*/

/*some attributes are abandoned, such as the time and place of counting becoming unimportant*/

protected:              /*"protected" indicates that the following behavior can be seen in a certain scope, such as the same
                          field*/

  int Counting( )          /*has a certain modularity in behavior, can take parameters and a return value*/
  {
      APP_List   app_list;    /*temporary variable for saving the apples set, but it is not sufficiently abstract*/

      Index(app_set)       /*ignore the shape of the apples set, and change them into a line when counting; shows that
                             counting does not relate to the arranged shape of objects*/
      {
         while(!app_set.Empty())
         {
            APPLE   an_apple; /*temporary variable shows that it is still in the field "apple-related"*/

            an_apple=app_set.SelectOneRandom();     /*mastered a counting principle: counting does not relate
                                                      to the arranged shape of objects*/

            app_list.Append(an_apple);

            app_set.Delete(an_apple);   /*mastered a counting principle: each object must be counted only once*/
         }
      }

      OneToOneMap()      / * Discovered the essential characteristic of counting: match an object with a numeral; it
                            is more advanced than a series of atomic operations*/
      {
         result=0;

         p.PointTo(app_list.First());

         p.Say(numlist.First());

         result++;

         while(app_list.Next()!=NULL)
         {
```

```
                p.PointTo(app_list.Next());

                p.Say(numlist.Next());

                result++;

            }

        }

        return    result;           /*shows that the behavior has been split into phases by logic*/

    }

}
```

Table 3 describes the typical behaviors for mastering the concept of "counting" in the Explicit E2 level, the inner characteristics of representation, and the usable strategy for formalization.

|   | Typical behaviors of Explicit E2 | Explanation of behaviors from inner representation | Strategy of formalization |
|---|---|---|---|
| 1 | The counting behavior of children achieved a new level; they can count several objects correctly, and the knowledge of counting can be applied to other domains easily. | The representation are more abstract, can be adapted to more problem solving, and the external visibility and shareability were improved. At this level, the representation can be called by other fields, and it also can call representations in other fields. | In formalization, the attributes of class become more abstract, and appear friend class, such as the class "numlist", which is used to represent natural numbers. Moreover, the external visibility of all operations becomes "public", allowing them to be used by the classes of other fields. |
| 2 | Children not only can count apples, but they can also count other objects. such as pencils, cups etc. The principle of object-unrelated is explicit. | The counting object is not restricted, and the constraints of the counting object are canceled. The new abstract representation can denote other kinds of objects. | The attribute used to denote the counting objects becomes object_set. This is an abstract data type showing that the children can count other kinds of objects. |
| 3 | Children's counting behavior becomes more flexible; for example, they can count objects without a fixed arranged sequence, as long as each object is counted only once. The principle of sequence-unrelated is explicit. | The representation is further reduced and the interference of accidental factors is ignored. It can utilize some attributes and operations of mature concepts, such as using the natural number concept or finding defined functional components through watching. It becomes more abstract and modularized. Thus, children can master the essence of concepts easily. | There are several independent functional modules appearing in the class, such as the function OneToOneMap() and the function GetResult(). Their coding is high mature, high cohesion and low coupling. |
| 4 | At this level, the children have the ability to "fetch objects by number", which is a new development of counting behavior. In addition, the children can solve other problems. For example, when you ask a child which figure is bigger between 5 and 7, the child will say "please bring some things and let me count". After "fetching objects by number", the child can make a correct judgment through comparing the sum of the | In representation, "natural number" is saved as an independent concept, and it becomes more specific. Some operations in the representation can be used by requirement | We use "numlist" to indicate the concept of natural number, and use "const" to indicate a constant. The constraint of attributes becomes "protected", and the constraint of operations becomes "public". Thus, they can be accessed by other classes, and can be used to solve new problems. |

| | two groups of things. It can be seen that the children have made new progress in comparing figures; this is a new skill based on the concept of counting. |
|---|---|

Table 3 Behavioral characteristics of explicit E2 and the strategy for realizing representation

```
const    intList   numlist;       /*natural number is separated and stored as independent knowledge*/
Class    Counting              /*do not care about the type of the counting object; closer to the natural essence of
                                   counting*/
{
    protected:        /*indicates that the following attributes can be seen in a certain scope, such as the same field*/
        friend     numlist;   /*friend class appears; all attributes can be shared by friend classes*/
        Person     p;
        objectSet    object_set;   /*the principle of object-unrelated appears*/
        Number    result;

    public:   /*totally explicit, and the following behavior can be shared by other fields*/
        void Index(object_set)
        {
            while(!object_set.Empty())
            {
                OBJECT an_object;
                an_object=object_set.SelectOneRandom();
                list.Append(an_object);
                object_set.Delete(an_object);
            {
        }
        void OneToOneMap(object_list) /*better modularity; the function "OneToOneMap" becomes independent*/
        {
            result=0;
            p.PointTo(object_list.First());
            p.Say(numlist.First());
            result++;
            while(object_list.Next()!=NULL)
```

```
            {
                p.Say(numlist.Next());
                result++;
            }
        }
        int GetResult( )
        {
            return    result;
        }
        int Counting( )
        {
            List    object_list;
            Index(object_set);
            OneToOneMap(object_list);
            GetResult( );
        }
    }
```

Table 4 describes the typical behaviors in mastering the concept of "counting" in the Explicit E3 level, the inner characteristics of representation, and the usable strategy for formalization.

|   | Typical behaviors of Explicit E3 | Explanation of behaviors from inner representation | Strategy of formalization |
|---|---|---|---|
| 1 | Children learnt the conservation of number; for example, let a child observe two heaps of candy, and the amount of each heap will become known by the child, but the amounts will have slight differences. Then we put the two heaps of candies into two similar transparent containers. When asked "Which container has more candies?", the child will quickly give the right answer, and explain that "the amount of candy did not change in this process". | The representation is divided into more detailed parts, and its modularity becomes more abstract, even appearing as a highly abstract concept, such as the concept "Set". The former, comparatively rough concept is divided into several more definite concepts, and the connotation of each concept becomes more detailed, single and accurate. Thus, the changes of concept attributes and the consequences of conceptual operations become very clear. | Representational form is divided into several classes, including two important classes: "counting" and "Set". The granularity and difficulty of representation decreases, but the relationship between concepts increases. The class "Set" has an important attribute, cardinalSum. Each Set has a special cardinal sum, and the cardinal sum is conservational. |
| 2 | In the task of comparing the amounts of two groups of objects, children will use the one-to-one matching method to compare them. They can make a correct judgment, namely, that the group with | Some operations have been used to perform the counting behavior, and now are completely explicit, such as the one-to-one matching operation. These operations can be used by the | We use the class "counting" to represent the concept "counting"; all attributes and operations of the class are public. The class includes the function OneToOne- |

| | surplus objects is has more than the other, and do not require the exact sum of the two groups of objects. | outside parts, and can allow the children to solve related problems with a new approach. | Map(), which can be used by other classes. |
|---|---|---|---|
| 3 | Children can use the one-to-one matching method to match objects, and they are not limited to matching objects with a number. For example, if there is a bus with 10 seats, and a child is asked "How many children can sit on the bus?", the child will first count the bus seats, and then get the correct result, according to OneToOne method. | The one-to-one matching method has been extended, and it can be adapted to more fields. At the same time, the concept "Set" can be embedded into other problem solving tasks as a parameter, because it is an abstract data type. | Two new functions appeared in the class "counting": OneToOne-Map(set1, set2) and Can_Match-_Discretely(set1, set2). They all relate to the mapping between two sets. Actually, they have a common core, but can solve different problems. Thus, they improved the applied flexibility of representation. |

Table 4 Behavioral characteristics of explicit E3 and the strategy for realizing representation

```
Class OrdinalNumber      /*separate or incorporate class according to the dependence between the components of
                            representation */
{
    private:
        const    intList    numlist;
        int       current;
        int       pre, succ;
    public:
        int       GetPre();
        int       GetNext();
        int       GetCurrent();
}

Class    Set     /*the counting behavior becomes a means to acquire the cardinal sum of a set*/
{
    public:            /*totally explicit, and the following attributes can be shared by other fields*/
        objectList       objlist;
        Boolean          item_type_be_similar=NO_OBLIGATORY;
        Boolean          item_sequence=NO_IMPORTANT;
        Boolean          item_arrangement=NO_IMPORTANT;
        int              cardinalSum; /*集合的基数*/
}

Class    Counting
```

```
{
    public:      /*totally explicit, and the following attributes can be shared by other fields*/
        Person         p;
        Set            set1, set2;
    public:
        int Counting();
        Boolean Can_Match_Discretely(set1, set2);
        int OneToOneMap(set1,set2);   /*the function "OneToOneMap" becomes independent and more general;
                                       it can be used for any two sets*/
}
```

## 5. The evolution of concept representation causes the improvement of problem solving ability

In our example, the representation of a concept developed from the Implicit I level to the Explicit E3 level in the process of concept acquisition in children. A rough psychological mechanism of information processing preserved the variation of representation. This is similar to the process of biological evolution, which uses genes to solidify some species characteristics. The development of a single concept representation promotes the ability of problem solving. In the above section, we used object-oriented formal specification to realize the process of RR; the outcome of concept acquisition was materialized by means of axiomatic semantics. The process of concept acquisition was gradual, and its representation increase one level after each RR. The form and content of the representation also improved. The ability of representation, the extent of modularity, and the scope of adaptation showed obvious development. Along with the concept representation becoming increasingly precise, children learned to apply the knowledge more flexibly, which cause their problem solving ability to continually evolve. Taking this description of children's counting as an example, we now introduce the evolution of the RR representational form and the evolution of problem solving ability.

**Implicit I:** There are several independent instances in representation, the constraints of all attributes and operations are "private", and all attributes are constant, with the behaviors of objects consisting of a series of atomic operations. The representation at this level is stiff, all attributes and operations can not be seen from outside, and they can not be shared. The presentation does not have any modularity, just some scenes. It is only adaptable to the same instance that already exists in the representation, such as the same row of apples.

**Explicit E1:** The instances become classes through inducing. The constraints of all operations become "protected". The attributes of class are represented by variables, the operations are represented by functions, and the functions are separated by their logical sense. The representation of this level has a certain abstraction, and it also has a certain level of parameter. Some operations also have a certain external visibility and can be used in the same field. In addition, the modularity becomes better, but it can only be used in limited fields. The representation of this level can be adapted to similar scenes, such as counting rows of apples with different sums. Children can solve some specific problems; for example, they have the ability of determining the sum of a row of apples.

**Explicit E2:** "numlist" becomes an independent constant. The constraints of attributes become "protected", and the constraints of operations become "public". At this level, the friend class and the functions OneToOneMap() and getResult() appear. Some attributes have a certain external visibility and even can be used by some objects in other fields, while some operations are totally visible and can be shared by other fields. In addition, the modularity can be used easily, and the logic components become clearer. The representation of this level can be adapted to many different scenes, and can even be applied to the problem solving of other fields, such as the counting of a set of pencils or cups. It also has the ability of to "fetch objects by number"[37,38]; for example, "I take five bananas" leads to the OneToOneMap method being called. The following is the formalized code of this process.

```
/ *new problem solving: "fetch objects by number", such as "I take five bananas", and OneToOneMap() is called*/
FetchObjects (Bananaset, 5)
{
    if(Counting.getResult()<5)
        ME.Say("Error");
    else
    {
        int i=0;
        while(i<5)
        {
            ME.TakeAway(Bananaset.SelectOneRandom());
            i++;
        }
    }
}
```

At this level, the ability "fetch objects by number" can be used to compare two numbers; for example, when you ask a child which figure is bigger between 5 and 7, and the child says "Please bring some things to let me count" [39], he will first fetch several objects, and then compare the sum of the two sets of objects, and then finally give the correct answer.

**Explicit E3:** At this level, three classes appear: OrdinalNumber, Set and Counting. The constraints of all attributes and operations in the three classes become "public", and the presentation becomes clearer. The big concept is separated into several small concepts, and the function of each module becomes more unique. All attributes and operations are totally visible, and thus they can be used by many other fields. Children can use this knowledge to solve problems in other fields, such as using the function OneToOneMap() to execute matching[38]. For example, if there is a bus with 10 seats, and a child is asked "How many children can sit on the bus?", the child will first count the seats of the bus, and then get the correct result according to OneToOneMap(). The formalized code of this process is as follows :

```
/*solving a more complex problem, such as: "How many passengers can sit on the bus"*/
Class    Set   Seats_of_Car
{
    cardinalSum=10;
}
Class Set Passengers
{
    cardinalSum=?;
}
if(Counting.Can_Match_Discretely(Seats_of_Car, Passengers))
        Passengers. cardinalSum=Seats_of_Car.cardinalSum;
else
{
        Counting.OneToOneMap(Seats_of_Car, Passengers);
        return Passengers.cardinalSum;
}
```

An additional problem is the conservation of number, such as when , for example, the arrangement of apples changes but this does not influence their sum. The formalized code of this process is the following :

```
/*solving a more complex problem, such as conservation of number*/
Class Set Apples
{
    cardinalSum=16;
    arrangement=SQUARE;
```

```
}
p.Move(Apples);
Apples.arrangement=CIRCLE;
Boolean CHANGE=ARRANGEMENT;
if(CHANGE==ARRANGEMENT)
      return Apples.cardinalSum;
else
      return Counting.Counting(Apples);
```

This demonstrates that the child acquired the OneToOneMap() through counting and could use the OneToOneMap() to solve a new problem.

## 6. Conclusion

The starting point of this research was the study of the development of conceptual systems from the algorithmic level, and to find a method to construct conceptual systems that can be used for solving unrestrictive problems. This paper takes previous research of cognition and conceptual representation in the fields of developmental psychology and cognitive psychology as reference, and in particular Karmiloff-Smith's RR hypothesis. This hypothesis put forward a multi-level representational method of constructing conceptual systems. It describes in detail the representation of each level by object-oriented means. On the whole, the current study has the following characteristics :

First, our research aim was to construct a conceptual system for the purpose of solving unrestrictive problems. This type of system is different from traditional expert systems, which deal only in one particular field. In this research, the form of representation is general, and as such it provides a foundation for solving unrestrictive problems in different areas.

Second, in this research the conceptual system has self-learning ability, namely, Representational Redescription, allowing the conceptual system to become more and more complete over time. The initial knowledge entries of the conceptual system is not sufficiently abstract enough; they are comprised of very specific materials. Abstract and formalized knowledge is then obtained by Representational Redescription. This design provides rich and detailed semantic support for, and enhances the completeness of, the conceptual system.

Third, taking the research of psychology as a reference is an obvious characteristic of this study. The main purpose of psychological research is to explore the cognitive structure of human beings. This stance is close to our starting point, and so psychology can provide some valuable views and assumptions. However, most psychological research make broad-brush assumptions, with the RR hypothesis used in this study being a prime example. The RR hypothesis also has many gaps, such as the lack of a detailed description of what factors redescription is based on, and how it actually happens, etc. As for the RR theory, we adopted its multi-level representation assumptions. We also designed a formalized representation and a generation algorithm for each level. In this sense, to a certain extent this paper makes up for some of the gaps of the RR hypothesis.

In summary, this research provides a new method of constructing a conceptual system based on RR. It also makes important contributions to solving unrestrictive problems. The research should be considered a knowledge acquisition module rather than a cognitive module, because it helps us to know more about the representational form and using the principles of the conceptual system in detail. However, the RR model includes four representational levels and three changing phases, and thus there must also be a complex underlying mechanism to control the RR process. Unfortunately such details are not included in the RR hypothesis. Therefore it is difficult to realize the algorithm of representational evolution; this is an important direction for future research.